\documentclass[letterpaper]{article} 
\usepackage{aaai2026}  
\usepackage{times}  
\usepackage{helvet}  
\usepackage{courier}  
\usepackage[hyphens]{url}  
\usepackage{graphicx} 
\usepackage{amsmath}
\usepackage{booktabs}
\usepackage{multirow}

\usepackage{natbib}  
\usepackage{caption} 
\usepackage{algorithm}
\usepackage{algorithmic}

\usepackage{newfloat}
\usepackage{listings}
\DeclareCaptionStyle{ruled}{labelfont=normalfont,labelsep=colon,strut=off} 
\floatstyle{ruled}
\newfloat{listing}{tb}{lst}{}
\floatname{listing}{Listing}
\title{Scientific Knowledge–Guided Machine Learning for Vessel Power Prediction: A Comparative Study}
\author {
    Orfeas Bourchas\textsuperscript{\rm 1},
    George Papalambrou\textsuperscript{\rm 1}
}
\affiliations {
    \textsuperscript{\rm 1} Laboratory of Marine Engineering N.T.U.A. \\
    orfeasbourchas@mail.ntua.gr, 
    george.papalambrou@lme.ntua.gr
}
\usepackage{amsmath}
\usepackage{booktabs}
\usepackage{multirow}
\usepackage{textcomp}

\begin{document}

\maketitle

\begin{abstract}
Accurate prediction of main engine power is essential for vessel performance optimization, fuel efficiency, and compliance with emission regulations. Conventional machine learning approaches, such as Support Vector Machines, variants of Artificial Neural Networks (ANNs), and tree-based methods like Random Forests, Extra Tree Regressors, and XGBoost, can capture nonlinearities but often struggle to respect the fundamental propeller law relationship between power and speed, resulting in poor extrapolation outside the training envelope. This study introduces a hybrid modeling framework that integrates physics-based knowledge from sea trials with data-driven residual learning. The baseline component, derived from calm-water power curves of the form $P=cV^n$, captures the dominant power–speed dependence, while another, nonlinear, regressor is then trained to predict the residual power, representing deviations caused by environmental and operational conditions. By constraining the machine learning task to residual corrections, the hybrid model simplifies learning, improves generalization, and ensures consistency with the underlying physics. In this study, an XGBoost, a simple Neural Network, and a Physics-Informed Neural Network (PINN) coupled with the baseline component were compared to identical models without the baseline component. Validation on in-service data demonstrates that the hybrid model consistently outperformed a pure data-driven baseline in sparse data regions while maintaining similar performance in populated ones. The proposed framework provides a practical and computationally efficient tool for vessel performance monitoring, with applications in weather routing, trim optimization, and energy efficiency planning. 
\end{abstract}


\section{Introduction}
\subsection{Classical power prediction modeling}
The maritime transportation sector is the backbone of international trade, but its contribution to global carbon emissions has led to increasingly stringent regulatory frameworks aimed at reducing the industry's carbon footprint \cite{imo2020fourth}. Addressing these environmental and economic pressures requires the optimization of vessel performance through accurate prediction of operational parameters, most critically, the main engine power consumption. Early approaches to power estimation relied on simple empirical regression of sea-trial curves or calm-water resistance models \cite{carlton2018marine}. These methods capture the dominant cubic relationship (propeller law) between speed and shaft power but fundamentally fail to account for complex environmental and operational deviations (e.g., wind, waves, hull fouling, or off-design drafts) encountered in real-world service. Classical machine learning models, such as Gaussian Processes and Support Vector Regression \cite{BASSAM2023114613, GKEREKOS2019106282} have been employed, demonstrating improved capability to model non-linear relationships compared to simple regression. However, these methods still struggle with the high dimensionality and volatility of vessel data. More recently, Artificial Neural Networks, including Long Short-Term Memory (LSTM) variants, have been widely adopted due to their universal approximation capability \cite{hornik1989multilayer}. Studies have shown that ANNs achieve high accuracy in predicting shaft power and fuel consumption \cite{guo_evaluating_2023, cai_diversity_2024, chen_short-term_2024, la_ferlita_framework_2024, zhang_deep_2024, mahmoodi_evaluating_2025}. Domain-augmented LSTMs have even shown more stable training characteristics \cite{cai_diversity_2024}. In the last few years, gradient-boosted decision trees, notably XGBoost \cite{Chen2016XGBoost}, have emerged as the state-of-the-art for tabular vessel performance datasets \cite{nguyen_application-oriented_2023, agand_fuel_2023, fan_comprehensive_2024, kiouvrekis_explainable_2025, wang_novel_2024}. These models consistently show competitive accuracy, often outperforming ANNs and classical methods in the prediction of vessel performance parameters due to their robustness to non-linearity and data noise. Despite their high in-sample accuracy, all purely data-driven models suffer from limited extrapolation ability. When applied outside the observed operational envelope (e.g., predicting power at speeds or drafts not seen during training), these models often produce predictions that violate the fundamental propeller-law scaling, thus becoming physically inconsistent and unreliable.

\subsection{Physics-Informed and Residual-Based Modeling}
The recognized shortcomings of purely data-driven models have spurred the development of methods that embed physical knowledge into the learning process. Physics-Informed Neural Networks (PINNs), formally introduced by Raissi et al. \cite{raissi2017physicsI, raissi2017physicsII}, have been proposed for vessel performance modeling to enforce the underlying physical laws \cite{LANG2024121877, bourchas_physics_2025}. While PINNs can enforce physical consistency, they are extremely hard to train and can struggle to capture steep gradients and discontinuities present in real-world operational data, filled with noisy measurements when applied in the forward problem. Hybrid Residual Strategies have emerged in the broader physics-informed learning literature as a promising alternative \cite{howard2023stackednetworksimprovephysicsinformed, 11036674}. These methods decompose the complex prediction task: a simple physics-based component models the dominant relationship, and a data-driven component learns the residual or deviation caused by complex, unaccounted-for factors.

\subsection{Our Contribution}
Inspired by the residual-hybrid strategy, a novel hybrid modeling framework for main-engine power prediction is proposed. The new framework effectively combines the stability of physics-based models with the predictive power of modern ML. The vessel’s sea-trial curves are utilized as a physically consistent baseline, which captures the dominant power-speed relationship, while a non-linear regressor is employed to learn the residual corrections—the difference between the observed power and the baseline—caused by environmental factors (e.g., waves, wind) and operational factors (e.g., trim, draft). By limiting the machine learning task to predicting only the residual corrections, our methodology significantly enhances extrapolation stability and physical consistency beyond conventional data-driven approaches. The main contributions of this work are:
\begin{itemize}
    \item The development of a hybrid modeling framework that overcomes traditional extrapolation limitations in vessel power prediction by leveraging a physics-informed baseline coupled with a non-linear regressor.
    \item A comparative study applying the proposed hybrid framework to XGBoost, a simple Neural Network, and a Physics-Informed Neural Network (PINN) is conducted, demonstrating the framework's versatility and efficacy across diverse modeling paradigms.
    \item  Validation using in-service data from real vessel operations provides a robust, computationally efficient tool for maritime optimization applications like weather routing and fuel efficiency planning.
\end{itemize}

\section{Methods}

\subsection{Main Engine Power Modeling}
Traditional machine learning approaches estimate the required main engine power by fitting a regression model directly to operational data. Formally, the predicted power can be expressed as:
{\footnotesize
\begin{equation}
\hat{P}(\bar{X}) = f(\bar{X})
\end{equation}}
Here, the function $f(\cdot)$ is a complex machine learning model and $\bar{X}$ represents the input parameters to the model, which represent the operating condition. Usually, these parameters include the vessel's speed through water (S.T.W.), the vessel's mean draft (T), the vessel's trim, and the wind speed and direction, as well as several others. Even though this approach has yielded promising results in the literature, its effectiveness is tied to the quality and quantity of the available data. However, in most cases, the quantity of available clean data is scarce as the hull's and propeller's performance deteriorates over time, either by the accumulation of fouling or damage to the coating and paints.

\subsection{Hybrid Main Engine Power Modeling}

To address the limitations of purely data-driven models, particularly their inability to confidently extrapolate beyond the training data envelope, a hybrid modeling framework is proposed. This approach decomposes the complex power prediction problem into two distinct components: a physics-based baseline model and a data-driven correction model. This separation ensures that the fundamental physical relationships are captured by a robust, simple model, while the machine learning component is tasked with learning only the deviations from this baseline. This decomposition allows the model to confidently and accurately extrapolate because the core physics are already accounted for.

\begin{figure}
\centering
\includegraphics[width=0.45\textwidth]{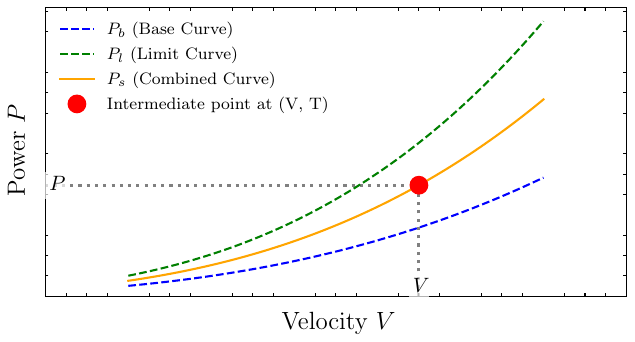}
\caption{Calm-water baseline construction. The laden ($P_l$) and ballast ($P_b$) sea-trial curves define the upper and lower bounds, while the combined curve $P_{\mathrm{sea\;trial}}(V,T)$ from Eq.~(3) provides the interpolated baseline at intermediate drafts. The red point marks the baseline value at the operating condition $(V,T)$.}

\label{fig:power_curves_plot}
\end{figure}

The baseline component is obtained from calm-water power curves measured at ballast and laden drafts during sea trials. These curves follow a power-law relation of the form
{\footnotesize\begin{equation}
P=cV^n
\end{equation}}
where $V$ is the vessel's speed through water, $c$ and $n$ are vessel-specific coefficients and are determined from the sea trial data, as presented in Appendix \ref{sec:Sea Trial}. For intermediate drafts as seen in Figure \ref{fig:power_curves_plot}, the baseline power is calculated by linear interpolation using equation \ref{intermediate_draft_power_prediction}:
{\footnotesize\begin{equation}
P_{\mathrm{sea\;trial}}(V, T) = \left(1 - \frac{T - T_b}{T_l - T_b}\right) \cdot P_b(V) + \left(\frac{T - T_b}{T_l - T_b}\right) \cdot P_l(V),
\label{intermediate_draft_power_prediction}
\end{equation}}
where $T_b$ and $T_l$ are the ballast and laden drafts, respectively, and $P_b(V)$ and $P_l(V)$ are the corresponding propeller power curves obtained from the sea trials. Even though the linear interpolation between drafts is a simplification of the underlying physics, it provides a robust baseline for intermediate drafts and satisfies the ballast and laden conditions simultaneously. The hybrid model's final prediction is the sum of this physics-based baseline and a learned correction term, $f(\bar{X})$, which accounts for all unaccounted factors such as weather, hull fouling, or engine degradation. 

\begin{figure}[t]
\centering
\includegraphics[width=0.45\textwidth]{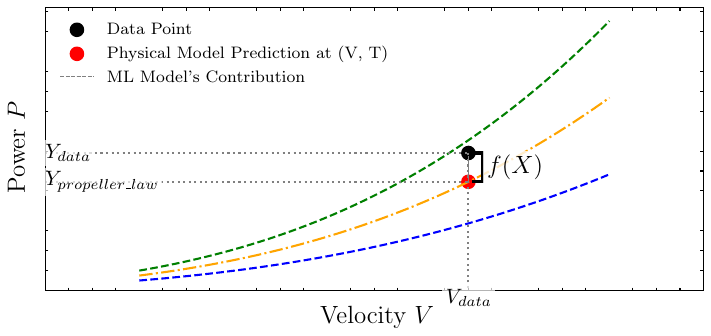}
\caption{Hybrid model decomposition as expressed in Eq.~(4). The laden and ballast sea-trial curves define the calm-water envelope, and the red point indicates the interpolated baseline at $(V,T)$. The black point represents an actual measurement. The vertical bracket labeled $f(\mathbf{X})$ illustrates the learned residual correction added to the baseline to obtain the hybrid prediction.}

\label{fig:residual_form_plot}
\end{figure}

The total predicted power, as seen in Figure \ref{fig:residual_form_plot} can be expressed as in  equation \ref{total_hybrid_power_prediction}
{\footnotesize\begin{equation}
    \hat{P}(\bar{X}) = P_{\mathrm{sea\;trial}}(V, T) + f(\bar{X})
\label{total_hybrid_power_prediction}
\end{equation}}

Here, $f(\cdot)$ is an XGBoost regressor, which is trained to predict the residual power. This residual is the difference between the measured main engine power and the sea trial baseline power. The input features to the machine learning model, $\bar{X}$, include parameters not captured by the sea trial data, such as wind speed and direction, wave height, vessel aging, and time since the last dry-docking. By constraining the machine learning model to learn only residuals, the prediction task becomes simpler and better regularized. This structure allows the model to generalize across a wider operational envelope, maintain consistency with physical laws, and extrapolate reliably beyond the training distribution.
\subsection{Deriving PINN's Loss}
The PINN is trained by minimizing a total loss that combines data fitting with a physics-based constraint.

The data loss is defined as
{\footnotesize
\begin{equation}
\label{eq:pinn_data_loss}
\mathcal{L}_{\text{data}} = \frac{1}{N} \sum_{i=1}^{N} \big( \hat{P}(\bar{X}_i) - P_i \big)^2,
\end{equation}}
and the physics loss $\mathcal{L}_{P.\text{law}}$ is given in Eq.~(\ref{eq_pinn_physics_loss}).  
The total PINN loss is then
{\footnotesize
\begin{equation}
\label{eq:eq_pinn_total_loss}
\mathcal{L}_{\text{PINN}}
= \mathcal{L}_{\text{data}} 
+ \lambda \, \mathcal{L}_{P.\text{law}},
\end{equation}}

where $\lambda$ is a weighting coefficient set to 100 in this work. Although fixed here for simplicity and stability, $\lambda$ can be treated as a tunable hyperparameter in future studies.

As described in \cite{bourchas_physics_2025}, the required power of a vessel's main engine $P$ with respect  to the vessel's speed through water $V$ and the operating condition is given by 
{\footnotesize\begin{equation}
\label{equation_5}
    P  \: =\:\frac{c_{T} \cdot W_{SA} \cdot V^3}{2 \cdot \eta_S \cdot \eta_{B} \cdot \eta_{H}} = \: c \cdot V^3
\end{equation}}
and the proposed physics loss is defined in Eq.~(\ref{eq_pinn_physics_loss})
{\footnotesize\begin{equation}
\label{eq_pinn_physics_loss}
\mathcal{L}_{P.\text{law}} \: =   \: \sum_{i=1}^{N} ( \frac{\partial {NN}}{\partial V}|_{x_{i}} - 3\cdot c\cdot V_{i}^2  ) ^2
\end{equation}}
where $P \simeq NN $. In the hybrid framework, $P$ is given by equation \ref{total_hybrid_power_prediction}. Taking the partial derivative of $P$ with respect to the vessel speed through water $V$:

{\footnotesize\begin{equation}
\label{equation_7}
\frac{\partial P}{\partial V} = \frac{\partial P_{\mathrm{sea\;trial}}(V, T)}{\partial V} +\frac{\partial f(\bar{X})}{\partial V} 
\end{equation}}
substituting equations \ref{intermediate_draft_power_prediction}, \ref{equation_5}, and rearranging the terms:

{\footnotesize\begin{equation}
\label{equation_8}
\begin{aligned}
\frac{\partial f(\bar{X})}{\partial V}  &= 3\cdot c\cdot V^2 \\
&-\left(1 - \frac{T - T_b}{T_l - T_b}\right) \cdot n_b \cdot c_b\cdot V^{n_b-1} \\
&- \left(\frac{T - T_b}{T_l - T_b}\right) \cdot n_l \cdot c_l\cdot V^{n_l-1}
\end{aligned}
\end{equation}}
where,
{\footnotesize\begin{equation}
\label{equation_9}
\begin{aligned}
\frac{\partial P_b(V)}{\partial V} &= n_b \cdot c_b\cdot V^{n_b-1},\\
\frac{\partial P_l(V)}{\partial V} &= n_l \cdot c_l\cdot V^{n_l-1},
\end{aligned}
\end{equation}}
All derivatives appearing in the physics loss are computed using \texttt{PyTorch}'s automatic differentiation engine (\texttt{autograd}).

\section{Dataset and Model Training}
\subsection{Dataset}
In the present work, the dataset from \cite{bourchas_physics_2025} was used. The dataset consists of approximately $40000$ data points, representing five months of operational data and is composed of the parameters described in Table \ref{tab:data_variables}. The dataset was randomly split into 80\%-10\%-10\% subsets representing training, validation, and testing sets. Drawing inspiration from the work of \cite{bourchas_physics_2025}, the \textit{Wind True Speed} and \textit{Wind True Direction} were decomposed into the $\textit{W}_\textit{x}$ and $\textit{W}_\textit{y}$ components using the equation \ref{eq:feature_engineering}

\begin{table}[t]
\caption{Dataset variables used as model inputs.}
\label{tab:data_variables}
\centering
\begin{tabular}{@{}l l c@{}}
\hline
Symbol & Description & Unit \\
\hline
$t$ & Time of the recording (UTC date) & -- \\
$P$ & Main engine brake power & kW \\
$V$ & Ship's speed through water (S.T.W.) & kn \\
$T$ & Ship's mean draft & m \\
$\mathrm{Trim}$ & Longitudinal trim & m \\
$\mathrm{WTS}$ & Wind true speed & kn \\
$\mathrm{WTD}$ & Wind true direction & ${}^{\circ}$ \\
\hline
\end{tabular}
\end{table}
{\footnotesize
\begin{equation}
\label{eq:feature_engineering}
\begin{aligned}
W_x &= \mathrm{WTS} \cdot \cos(\mathrm{WTD}), \\
W_y &= \mathrm{WTS} \cdot \sin(\mathrm{WTD}),
\end{aligned}
\end{equation}}

\subsection{Model Training and Hyperparameter Optimization}
To ensure a robust and fair evaluation, both the baseline and hybrid configurations of all three model architectures—XGBoost, Neural Network (NN), and Physics-Informed Neural Network (PINN)—underwent a unified and rigorous hyperparameter optimization (HPO) procedure.

For the XGBoost implementation, the widely used \texttt{XGBoost} package was used, and hyperparameter tuning was carried out using \texttt{Scikit-learn}'s \texttt{RandomizedSearchCV}. The optimization explored parameter distributions including the learning rate ($\eta$) in ${0.01, 0.05, \dots, 0.30}$, maximum tree depth in ${3, \dots, 10}$, number of estimators (\texttt{n\_estimators}) in ${5, \dots, 500}$, and L1 ($\alpha$) and L2 ($\lambda$) regularization coefficients in ${0, 1, 100}$. The negative mean squared error was used as the scoring metric. During training, both input and output variables were standardized using \texttt{Scikit-learn}'s \texttt{StandardScaler}, fitted on the training dataset.

The NN and PINN models were developed and trained within the \texttt{PyTorch} ecosystem, with hyperparameter optimization conducted via \texttt{Weights \& Biases (WandB)}'s \texttt{sweep} functionality. A Bayesian optimization strategy was adopted, minimizing the root mean squared error (RMSE) on the test set as the objective metric. The search space for both NN and PINN models was centered around the optimal settings suggested in \cite{bourchas_physics_2025}, exploring key parameters such as \texttt{learning rates} ${1\times10^{-2}, 1\times10^{-3}, 3\times10^{-4}, 1\times10^{-4}}$, \texttt{layer widths} of ${64, 128, 256}$ neurons, and \texttt{network depths} of ${4, 6, 8}$ layers. Each model was trained using the Adam optimizer with the selected learning rate for 1000 epochs, without learning rate scheduling or early stopping. For the PINN model specifically, the total loss was defined as in Eq.~(\ref{eq:eq_pinn_total_loss}), with $\lambda$ fixed to 100 as discussed in Section~2.3.

The final optimized hyperparameter values for all models (XGBoost, NN, and PINN), obtained through these comprehensive optimization procedures, are summarized in Tables \ref{tab:xgb_hpo}, \ref{tab:nn_hpo}, \ref{tab:pinn_hpo}.
\begin{table}[t!]
\centering
\caption{Optimized hyperparameters for XGBoost models (Baseline vs. Hybrid).}
\begin{tabular}{lcc}
\hline
\textbf{Hyperparameter} & \textbf{Baseline} & \textbf{Hybrid} \\
\hline
Learning rate ($\eta$) & 0.25 & 0.1 \\
Max depth & 10 & 10 \\
$n$-estimators & 40 & 500 \\
L1 regularization ($\alpha$) & 1 & 1 \\
L2 regularization ($\lambda$) & 0 & 100 \\
\hline
\end{tabular}
\label{tab:xgb_hpo}
\end{table}
\begin{table}[t!]
\centering
\caption{Optimized hyperparameters for Neural Network (NN) models (Baseline vs. Hybrid).}
\begin{tabular}{lcc}
\hline
\textbf{Hyperparameter} & \textbf{Baseline} & \textbf{Hybrid} \\
\hline
Learning rate & $1\times10^{-4}$ & $3\times10^{-4}$ \\
Hidden layers & 6 & 4 \\
Neurons per layer & 256 & 64 \\
\hline
\end{tabular}
\label{tab:nn_hpo}
\end{table}
\begin{table}[t!]
\centering
\caption{Optimized hyperparameters for Physics-Informed Neural Network (PINN) models (Baseline vs. Hybrid).}
\begin{tabular}{lcc}
\hline
\textbf{Hyperparameter} & \textbf{Baseline} & \textbf{Hybrid} \\
\hline
Learning rate & $1\times10^{-4}$ & $3\times10^{-4}$ \\
Hidden layers & 8 & 6 \\
Neurons per layer & 256 & 256 \\
\hline
\end{tabular}
\label{tab:pinn_hpo}
\end{table}

\subsection{Hardware Configuration}
All computational experiments were conducted on a machine with Windows 10 Pro as the operating system, an AMD Ryzen 5500 CPU, an NVIDIA RTX 3060 GPU with 12 GB of VRAM, and 32 GB of RAM. 

\section{Results and Model Evaluation}
This section evaluates the baseline and hybrid variants of the XGBoost, NN, and PINN models using both standard error metrics and a detailed analysis of their extrapolation behavior. Although the quantitative differences between baseline and hybrid models are modest, the qualitative analysis reveals clear advantages for the hybrid formulations in terms of physical consistency and robustness under unseen operating conditions.

\subsection{Quantitative Analysis}

Predictive accuracy was assessed using two standard metrics: Mean Absolute Error (MAE) and Root Mean Square Error (RMSE). After hyperparameter optimization, each model was evaluated on the training, validation, and test sets, with results summarized in Table~\ref{tab:performance_metrics_all}.

Overall, the XGBoost baseline achieved slightly lower training errors than its hybrid counterpart, which is consistent with its strong capacity to fit the available data. Similar behavior was observed for the NN and PINN architectures, where baseline models occasionally achieved marginally lower MAE and RMSE values. However, across all architectures, these quantitative differences remained small and well within acceptable limits for operational maritime applications.

It is important to note that such global error metrics do not capture model behavior outside densely populated regions of the training dataset. As shown in the next subsection, baseline models display clear degradation in extrapolation scenarios, whereas the hybrid variants consistently deliver more stable and physically coherent predictions.

\begin{table*}[t!]
\centering
\caption{Performance comparison across all architectures.}
\begin{tabular}{lcccccc}
\toprule
\multirow{2}{*}{\textbf{Metric}} &
\multicolumn{2}{c}{\textbf{XGBoost}} &
\multicolumn{2}{c}{\textbf{Neural Network (NN)}} &
\multicolumn{2}{c}{\textbf{Physics-Informed NN (PINN)}} \\
\cmidrule(lr){2-3} \cmidrule(lr){4-5} \cmidrule(lr){6-7}
 & \textbf{Baseline} & \textbf{Hybrid} 
 & \textbf{Baseline} & \textbf{Hybrid} 
 & \textbf{Baseline} & \textbf{Hybrid} \\
\midrule
\textbf{Train MAE [kW]}  
  & 114.2 & 143.3 & 181.01 & 235.30 & 147.03 & 214.73 \\
\textbf{Train RMSE [kW]} 
  & 180.5 & 199.9 & 229.67 & 292.12 & 205.48 & 248.30 \\
\midrule
\textbf{Validation MAE [kW]}  
  & 114.2 & 143.3 & 156.29 & 214.08 & 135.52 & 167.17 \\
\textbf{Validation RMSE [kW]} 
  & 180.5 & 199.9 & 212.94 & 272.38 & 204.21 & 219.51 \\
\midrule
\textbf{Test MAE [kW]}   
  & 122.2 & 148.8 & 162.66 & 219.32 & 144.30 & 171.19 \\
\textbf{Test RMSE [kW]}  
  & 195.0 & 208.2 & 225.10 & 284.33 & 211.89 & 229.45 \\
\bottomrule
\end{tabular}
\label{tab:performance_metrics_all}
\end{table*}

\subsection{Qualitative Extrapolation Analysis}

\begin{figure}[t!]
\centering
\includegraphics[width=0.85\linewidth]{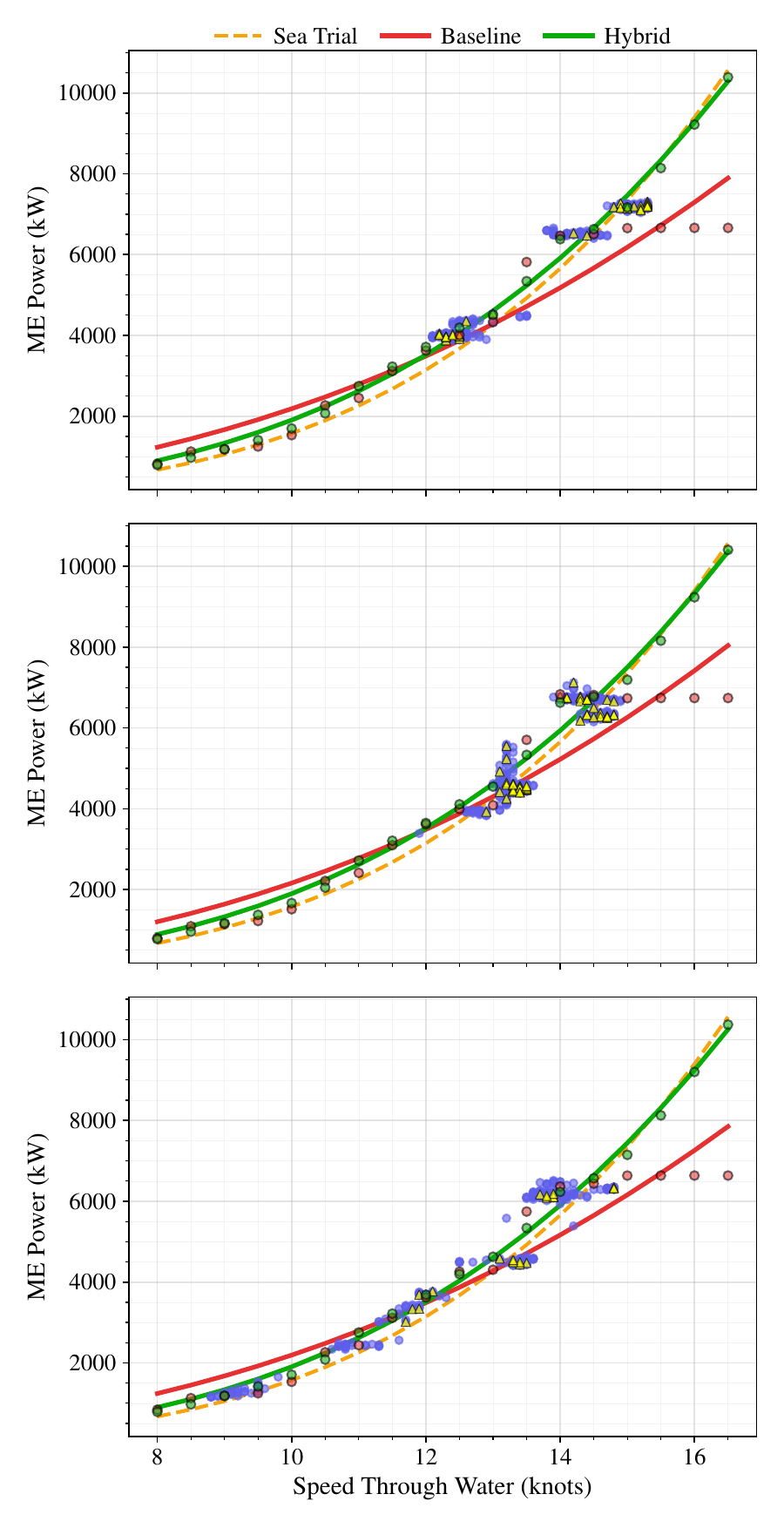}
\caption{Extrapolation behavior of the XGBoost baseline and hybrid models at ballast draft and 5-kn wind. Rows correspond to wind directions of $0^{\circ}$, $90^{\circ}$, and $180^{\circ}$. Red and green curves show the baseline and hybrid predictions, respectively. Blue points denote the nearest training data, while yellow triangles indicate nearest test samples.}
\label{fig:XGB_Ballast_Condition}
\end{figure}

\begin{figure}[t!]
\centering
\includegraphics[width=0.85\linewidth]{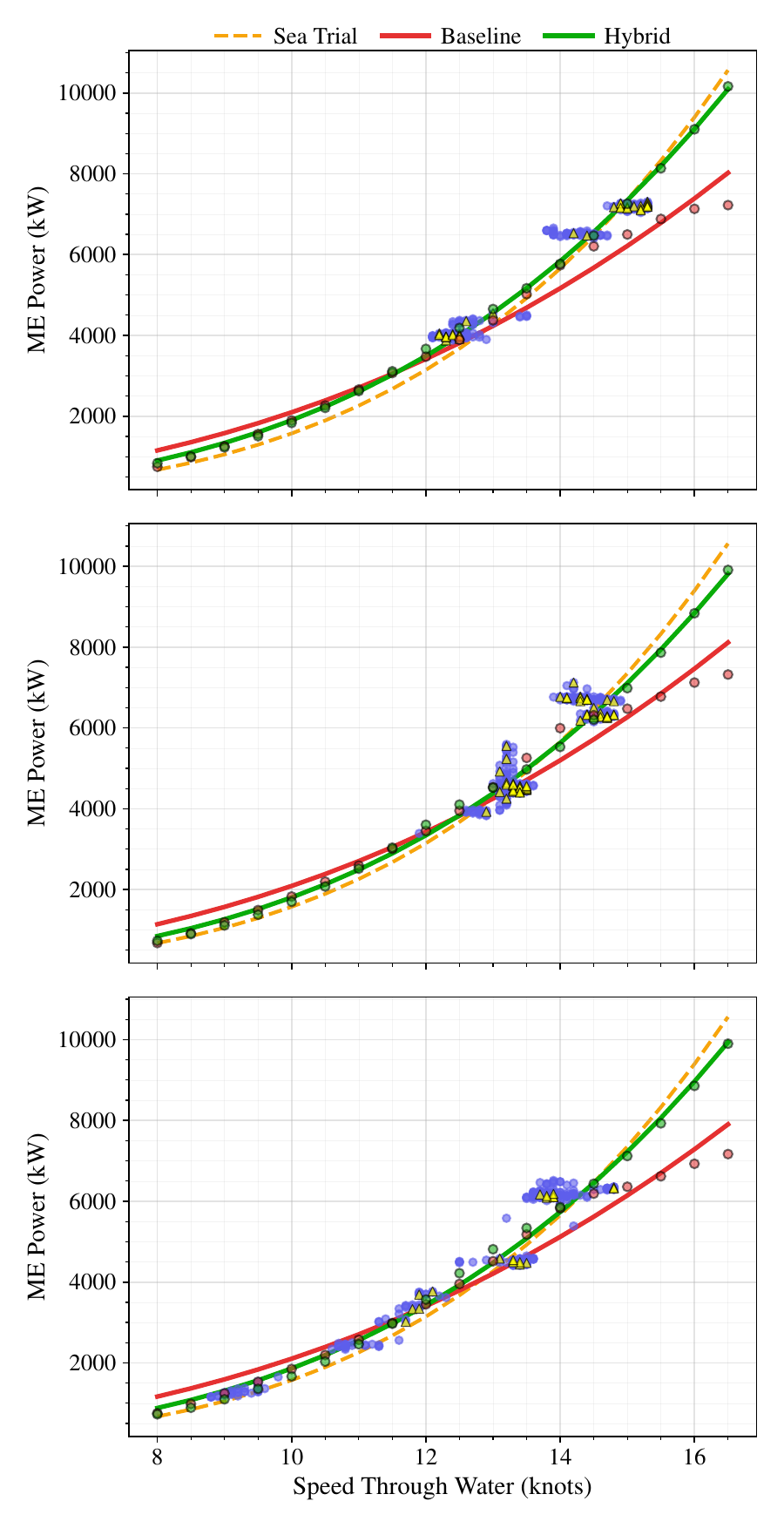}
\caption{Extrapolation behavior of the NN baseline and hybrid models at ballast draft and 5-kn wind. Rows correspond to wind directions of $0^{\circ}$, $90^{\circ}$, and $180^{\circ}$. Red and green curves show the baseline and hybrid predictions, respectively. Blue points denote the nearest training data, while yellow triangles indicate nearest test samples.}
\label{fig:NN_Ballast_Condition}
\end{figure}

\begin{figure}[t!]
\centering
\includegraphics[width=0.85\linewidth]{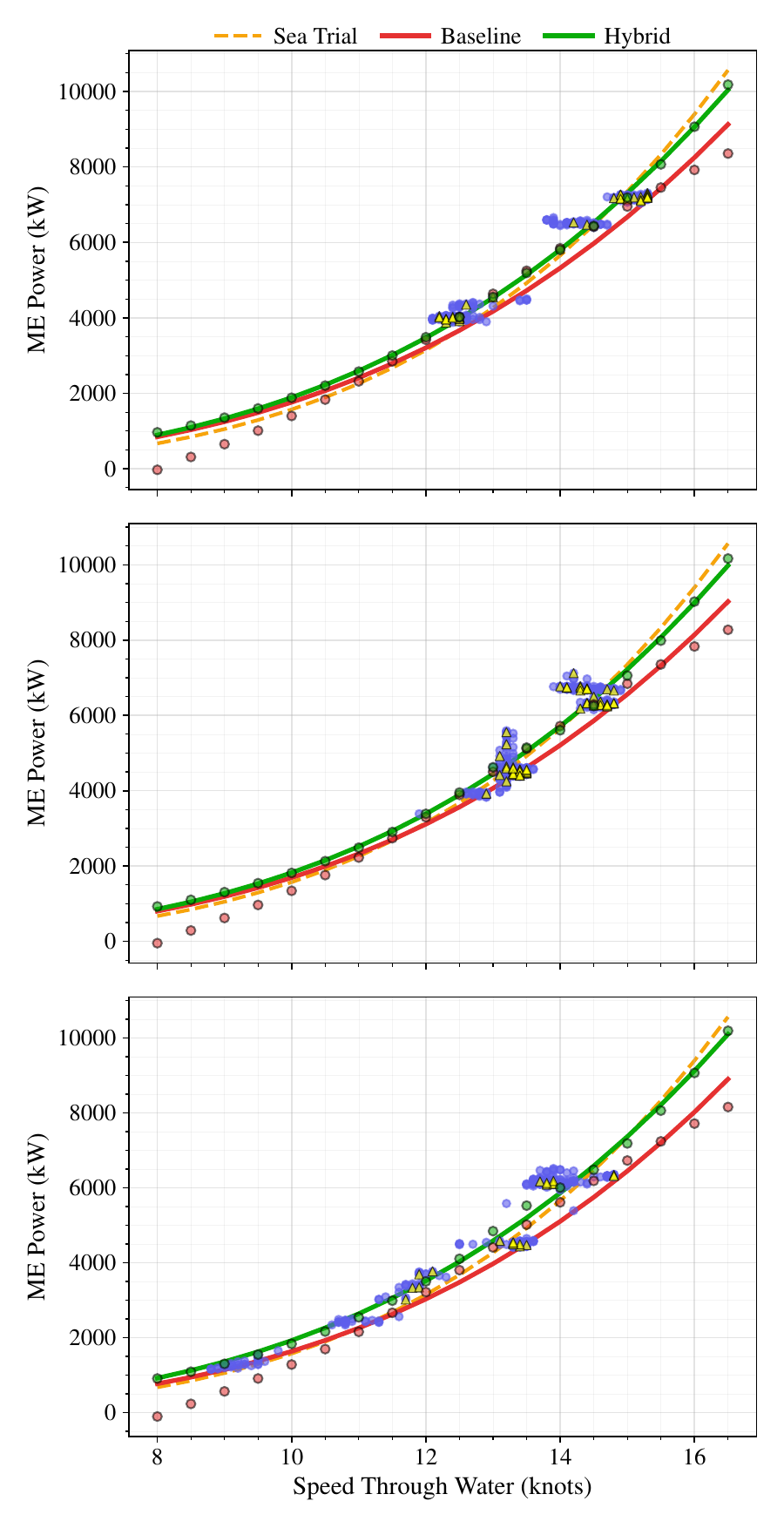}
\caption{Extrapolation behavior of the PINN baseline and hybrid models at ballast draft and 5-kn wind. Rows correspond to wind directions of $0^{\circ}$, $90^{\circ}$, and $180^{\circ}$. Red and green curves show the baseline and hybrid predictions, respectively. Blue points denote the nearest training data, while yellow triangles indicate nearest test samples.}
\label{fig:PINN_Ballast_Condition}
\end{figure}

To assess robustness under unseen operating conditions, each model family was evaluated across speeds from 8--17 kn at ballast draft and 5-kn wind, with wind directions of $0^{\circ}$, $90^{\circ}$, and $180^{\circ}$. Figures~\ref{fig:XGB_Ballast_Condition}, \ref{fig:NN_Ballast_Condition}, and \ref{fig:PINN_Ballast_Condition} present representative results for XGBoost, NN, and PINN, respectively. Blue points denote the closest training samples, while yellow triangles represent nearby test samples, highlighting regions where the models must extrapolate.

\paragraph{XGBoost (Figure~\ref{fig:XGB_Ballast_Condition}).}
The baseline XGBoost model frequently exhibits non-monotonic or flattened power–speed behavior, especially at higher speeds where training data are not available. Such distortions violate the physically expected propeller-law scaling and lead to predictions that are not operationally credible. The hybrid XGBoost model corrects this issue by anchoring the prediction to the calm-water baseline, which preserves a smooth and monotonic increase in power even where training data are sparse.

\paragraph{Neural Network (Figure~\ref{fig:NN_Ballast_Condition}).}
The baseline NN displays similar extrapolation deficiencies, including overestimation at high speeds and noticeable sensitivity to wind direction in poorly sampled regions. The hybrid NN significantly reduces these artifacts. Because it learns only residual deviations around a physically consistent baseline, the resulting power–speed curves are smoother and exhibit improved generalization across the full speed range.

\paragraph{Physics-Informed Neural Network (Figure~\ref{fig:PINN_Ballast_Condition}).}
The hybrid PINN demonstrates the strongest extrapolation performance across all architectures. While the baseline PINN benefits from its physics-informed loss, it can still deviate in sparsely represented regions. The hybrid formulation enforces consistency with both the sea-trial power law and the PINN derivative constraints, producing physically coherent predictions across all wind directions and throughout the 8–17 kn range.

\paragraph{Summary.}
Across all model families, the hybrid formulations consistently improve extrapolation stability while maintaining accuracy in dense data regions. These gains stem directly from embedding domain knowledge: by relieving the models from rediscovering the dominant power–speed law, the hybrid structure yields more reliable predictions in unseen or sparsely populated operating regimes. Among all architectures, the hybrid PINN demonstrates the best balance between numerical accuracy and physical consistency, making it particularly attractive for decision-support applications where extrapolation reliability is critical.

Taken together, these findings indicate that the main benefit of the hybrid formulations is not a reduction in error metrics, but a regularization of model behavior in sparsely sampled and extrapolative regimes. In the most demanding operating conditions examined, such as high speeds combined with adverse winds, the hybrid models remain close to the physically expected power–speed trend, whereas the baseline models may flatten or deviate from the calm-water envelope. This distinction is particularly important for applications like weather routing and speed optimization, where decisions are often made precisely in regions with limited historical data coverage.

\section{Conclusion}
Across all model families, the baseline and hybrid variants showed small quantitative differences, well under 1\% of the vessel’s maximum power, indicating comparable accuracy in standard error metrics. However, these metrics do not capture a critical distinction: extrapolation performance. Hybrid models consistently produced smoother and more physically coherent power–speed curves outside the training envelope, while baseline models, especially XGBoost, often exhibited non-physical behavior at higher speeds or under extreme environmental conditions. Among all architectures, the hybrid PINN delivered the best overall performance, combining competitive accuracy with the strongest physical consistency in unseen regimes. These results demonstrate that incorporating physical structure through hybridization significantly improves reliability under extrapolation. The proposed framework is well suited for real-world maritime decision-support applications, such as weather routing and trim optimization, where accurate power prediction is required even when no historical data are available.

\appendix
\section{Sea Trial Parameters}
\label{sec:Sea Trial}
Sea trials are a set of standardized tests used to determine a vessel's performance characteristics under controlled, calm weather conditions. During these trials, a ship's speed ($V$) and main engine power ($P$) are systematically measured at specific draft conditions, typically in ballast ($T_b$) and design or laden ($T_l$) states.

The relationship between power and speed is commonly approximated using a power law:

$$P = cV^n$$

where $c$ and $n$ are coefficients specific to the vessel. To determine these coefficients, the power law is linearized by taking the natural logarithm of both sides:

$$\ln(P) = \ln(c) + n \ln(V)$$

This equation represents a linear relationship between $\ln(P)$ and $\ln(V)$. By performing a linear regression on the log-transformed sea trial data, the coefficients $\ln(c)$ and $n$ can be determined, providing a robust power-speed curve for the vessel in calm water.

\bibliography{aaai2026}

\end{document}